\documentclass{article}
\usepackage{spconf,amsmath,graphicx}
\usepackage{amsfonts, multirow, booktabs}
\usepackage{threeparttable}
\usepackage{cite}
\title{FEW-SHOT ONE-CLASS DOMAIN ADAPTATION BASED ON FREQUENCY FOR IRIS PRESENTATION ATTACK DETECTION}
\name{Yachun Li, Ying Lian, Jingjing Wang, Yuhui Chen, Chunmao Wang, Shiliang Pu\sthanks{Corresponding author (pushiliang.hri@hikvision.com).}}
\address{Hikvision Research Institute}

\begin{document}
\maketitle
\begin{abstract}
Iris presentation attack detection (PAD) has achieved remarkable success to ensure the reliability and security of iris recognition systems. Most existing methods exploit discriminative features in the spatial domain and report outstanding performance under intra-dataset settings. However, the degradation of performance is inevitable under cross-dataset settings, suffering from domain shift. In consideration of real-world applications, a small number of bonafide samples are easily accessible. We thus define a new domain adaptation setting called Few-shot One-class Domain Adaptation (FODA), where adaptation only relies on a limited number of target bonafide samples. To address this problem, we propose a novel FODA framework based on the expressive power of frequency information. Specifically, our method integrates frequency-related information through two proposed modules. Frequency-based Attention Module (FAM) aggregates frequency information into spatial attention and explicitly emphasizes high-frequency fine-grained features. Frequency Mixing Module (FMM) mixes certain frequency components to generate large-scale target-style samples for adaptation with limited target bonafide samples. Extensive experiments on LivDet-Iris 2017 dataset demonstrate the proposed method achieves state-of-the-art or competitive performance under both cross-dataset and intra-dataset settings.

\end{abstract}

\section{Introduction}
\label{sec:intro}

Iris has unique and abundant texture information and has been widely used in biometric recognition applications of high-security levels. However, iris recognition system is vulnerable to various presentation attacks, causing security concerns in our society. 

To alleviate these issues, presentation attack detection (PAD) has drawn growing attention. As deep learning has proved its ability in many applications, many methods employ CNNs to develop PAD systems to detect spoof samples\cite{chen2018multi, kimura2020cnn, he2016multi, raghavendra2017contlensnet, fang2020micro}. 
Nevertheless, these methods exploit discriminative features for iris PAD only in the spatial domain. Bonafide iris has rich textures, while attacks such as the colored contact lenses have specific patterns and printouts would show some artifacts due to print quality. These differences can be easily captured in the frequency domain and are rarely exploited in iris PAD yet. 
Besides, frequency information has demonstrated its power in many related fields. It has been seamlessly inserted in networks to incorporate frequency components to color space\cite{xu2020learning, qin2021fcanet}. Additional frequency domain analysis explores spoofing clues in both face liveness detection\cite{kim2012face, fang2021learnable} and forgery detection\cite{chen2021local, qian2020thinking} tasks. 

Therefore, we propose a novel Frequency-based Attention Module (FAM). 
FAM aggregates frequency information into spatial attention and is embedded in multiple layers to explicitly highlight high frequency components. Specifically, we employ Discrete Cosine Transform (DCT) to obtain frequency-based features. Then a learnable mask is applied to filter out low frequency components and keep high frequency ones. The masked frequency-based features are finally converted back to spatial domain through inverse DCT and act as the frequency-based attention to original features. FAM is different from SE\cite{hu2018squeeze} and CBAM\cite{woo2018cbam} since they only focus on spatial domain. The most related work is FcaNet\cite{qin2021fcanet}. The similarity is that we both integrate frequency information into the attention map. However, FcaNet presents multi-spectral channel attention aiming at more efficient channel representation, while our FAM generates spatial attention in multiple layers to highlight high frequency components for iris PAD.

\begin{figure*}[t]
\begin{center}
\includegraphics[width=1.\linewidth]{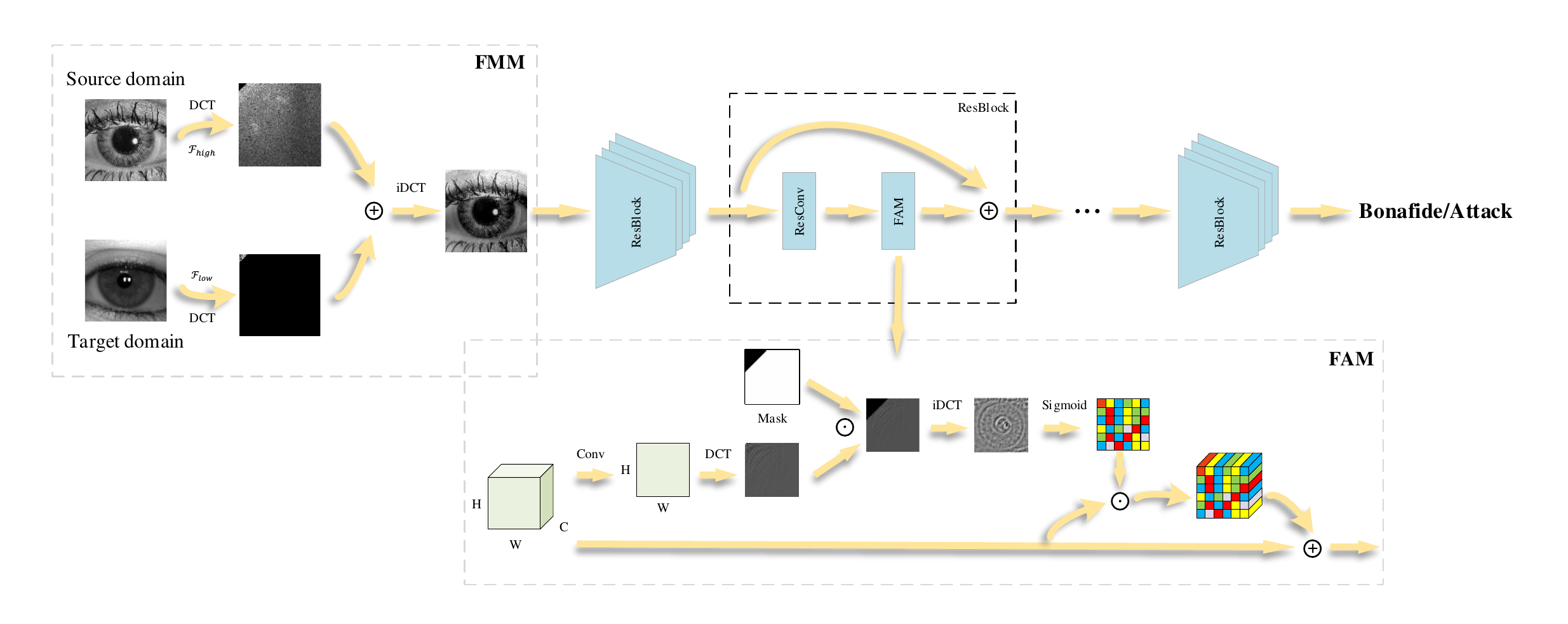}
\end{center}
\vspace{-1cm}
\caption{Overview of few-shot one-class domain adaptation framework. Frequency information is leveraged in both frequency mixing module and frequency-based attention module. }
\label{fig:overview}
\vspace{-0.2cm}
\end{figure*}

Although current iris PAD works achieve promising results in intra-dataset settings, the performance inevitably degrades under cross-dataset testing, due to domain shift. To alleviate this, domain adaptation (DA) techniques have been introduced to face anti-spoofing\cite{li2018unsupervised, wang2021self}. It has not been employed in iris PAD field yet. Moreover, face anti-spoofing DA methods utilize a large number of unlabeled images, including both live and spoof ones. It is of high cost to collect attack samples, but collecting several bonafide samples from target domain should be convenient. Therefore, we define a new few-shot domain adaptation setting called Few-shot One-class Domain Adaptation (FODA), where only a small number of bonafide samples from target domain are available.

Inspired by \cite{yang2020fda}, we propose a new Frequency Mixing Module (FMM), which generates large-scale target-style samples with limited target bonafide samples by mixing certain frequency components. To be specific, low frequency components are more style-related and contribute more to domain shift, while high frequency components contain more bonafide/attack-related contents. We mix low frequency components of images from target domain with high frequency components of images from source domain, in order to generate new samples with source labels and target styles. 

Finally, our proposed FODA framework combines FAM and FMM. FMM first generates target-style samples with few target bonafide samples. Original and adapted source samples are fed to the FAM-embedded network. Multi-scale FAM helps to explicitly emphasize high frequency components and benefits PAD-related feature learning. Two frequency-based modules integrate efficient frequency information and facilitate iris presentation attack detection.

\section{Proposed Method}
\label{sec:method}

Fig. \ref{fig:overview} shows an overview of our few-shot one-class domain adaptation framework, including two modules. Frequency-based Attention Module (FAM) adaptively decouples PAD-related high frequency features. High frequency components are extracted from features through discrete cosine transform (DCT), and then converted back to spatial domain via inverse DCT to get frequency-based attention. Frequency Mixing Module (FMM) is designed to address the domain shift problem. It generates a large number of target-style samples with very few target bonafide samples. By mixing certain frequency components of source and target images, the mixing images have source labels and target style at the same time. This helps to reduce the domain shift caused by style discrepancy. Both source and mixed samples are fed to the network, and the multi-scale FAM network explicitly emphasizes high frequency components to facilitate PAD feature learning.
We describe the two frequency-based modules in detail next.

% Ablation
\begin{table*}[t!]
 \begin{center}
\begin{threeparttable}
\footnotesize
\caption{Ablation studies under cross-dataset settings. }
\label{tab:ablation}
  \begin{tabular}{c|c|cc|cc|cc|c}
  \toprule[2pt]
  \multicolumn{2}{c|}{Trained Dataset}  & \multicolumn{2}{c|}{IIITD-WVU} & \multicolumn{2}{c|}{NotreDame} & \multicolumn{2}{c|}{Clarkson}   & \multirow{2}{*}{Average} \\ \cline{1-8}
  \multicolumn{2}{c|}{Tested Dataset}   & NotreDame     & Clarkson       & IIITD-WVU      & Clarkson      & IIITD-WVU      & NotreDame      &                          \\
  \midrule[1.3pt]
  \multirow{4}{*}{wo/DA} & Baseline     & 7.33          & 45.69          & 20.97          & 11.23         & 29.46          & 28.83          & 23.92                    \\
  & SE           & 5.75          & 42.26          & 19.17          & 13.33         & 31.47          & 34.14          & 24.35                    \\
  & CBAM         & 9.11          & 30.72          & 13.47 & 14.08         & 36.78          & 31.17          & 22.56                    \\
  & FAM          & 12.86         & 28.90           & 14.38          & 9.39          & 25.88          & 19.81          & 18.54                    \\ \hline
  \multirow{6}{*}{w/DA}  & DANN\tnote{*}         & 10.53         & 25.57          & 14.71          & 20.45         & 27.83          & 19.94          & 19.84                    \\
  & MMD\tnote{*}          & 20.31         & 40.58          & \textbf{11.36}          & 23.31         & 26.39          & \textbf{14.14} & 22.68                    \\
  & FMM      & 4.50           & 36.31          & 18.49          & \textbf{8.91} & 22.83          & 24.61          & 19.27                    \\
  & FMM+SE   & \textbf{3.83} & 27.72          & 21.47          & 14.72         & 29.03          & 29.72          & 21.08                    \\
  & FMM+CBAM & 6.36          & \textbf{24.99} & 14.76          & 11.82         & 36.85          & 35.53          & 21.72                    \\
  & FMM+FAM  & 5.81          & 26.03          & 15.07          & 10.51         & \textbf{22.06} & 20.92          & \textbf{16.73}                    \\ 
  \bottomrule[2pt]
 \end{tabular}
\begin{tablenotes}\footnotesize
\item[*] Entire unlabeled target domain images were used for adaptation.
\end{tablenotes}
\end{threeparttable}
 \end{center}
\vspace{-0.8cm}
\end{table*}

\subsection{Frequency-based Attention Module}
\label{ssec:att}
Images can be decomposed into low and high frequency components. Low frequency components are usually related to style content, while high frequency components contain more fine-grained information. As iris PAD task is more concerned with subtle textures, emphasis on high frequency ones could bring more generalizability to the model. To this end, we propose to explicitly enhance high frequency part via FAM. 

Specifically, given a feature map $\mathbf{f} \in \mathbb{R}^{C \times H \times W}$, FAM first aggregates multiple channels to single one, where the output feature $\mathbf{f}_{a} \in \mathbb{R}^{H \times W}$ is calculated by:

\vspace{-0.3cm}
\begin{equation}\label{aggre}
\mathbf{f}_{a} = F_{aggre}(\mathbf{f}) = \sum_{i=1}^{C}  W_i \cdot \mathbf{f}_i
\vspace{-0.1cm}
\end{equation}

Discrete Cosine Transform (DCT) is applied to one-channel feature $\mathbf{f}_{a}$. And then we leverage a learnable mask $\mathbf{M}$, initialized to remove the low-frequency band, in order to adaptively filter out low frequency components while keep high frequency ones in frequency domain: 

\vspace{-0.3cm}
 \begin{equation}\label{dct}
\mathbf{f}_{highfreq} =  \mathcal{D}(\mathbf{f}_{a}) \odot \mathbf{M}
\vspace{-0.1cm}
\end{equation}
where $\mathcal{D}$ is DCT and $\odot$ is the element-wise product. 

Finally, we employ inverse DCT to transform high frequency feature to spatial domain and scale the original feature $\mathbf{f}$ with this frequency-aware attention map. The frequency-based attention is obtained by:

\vspace{-0.3cm}
 \begin{equation}\label{att}
\mathbf{f}_{freqatt} =  \sigma(\mathcal{D}^{-1}(\mathbf{f}_{highfreq}))
\vspace{-0.1cm}
\end{equation}
where sigmoid function $\sigma(x) = \frac{1}{1+e^{-x}}$ aims at squeezing $x$ within the range $(0,1)$. We adopt a residual connection to get the final frequency-enhanced feature $\mathbf{f}_{FAM} = \mathbf{f} + \mathbf{f}_{freqatt}  \odot \mathbf{f}$. Thus, high frequency components are explicitly emphasized through the proposed frequency-aware attention mechanism.

\begin{figure}[t!]
\begin{center}
\includegraphics[width=1.\linewidth]{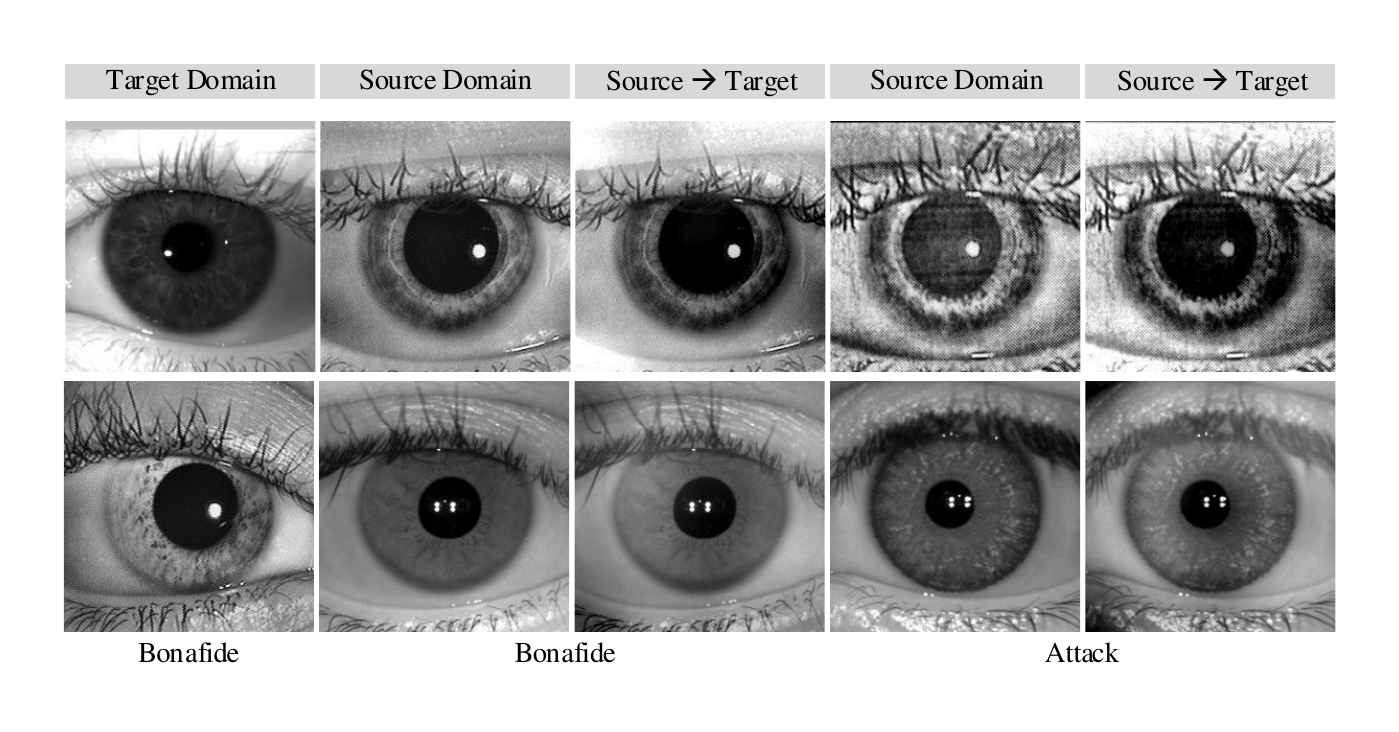}
\end{center}
\vspace{-1cm}
\caption{Mixed samples from FMM. }
\label{fig:example}
\vspace{-0.5cm}
\end{figure}

\subsection{Frequency Mixing Module}
\label{ssec:mix}
Considering domain adaptation scenarios, there are distribution biases between training and testing domains. The discrepancy of domains might be from capturing equipment, illuminations, backgrounds, etc. Therefore, we would like to generate samples that are able to imitate the possible changes.
The high frequency components contain textures features that are crucial to detect presentation attacks. 
Domain style is more related to low-frequency components, which may help to distinguish different domains. Based on this, we present a novel module called FMM to adapt model towards target domain with few costs. 

For presentation attack detection task, bonafide samples are more accessible than attack ones. We thus define the typical domain adaptation problem, where only a small number of bonafide samples from target domain are available (e.g. 10 bonafide samples). 
Given source domain $\mathcal{S}=\{ (x^s_i, y^s_i) \}_{i=1}^{N_s}$ and target domain $\mathcal{T}=\{ (x^t_i, y^t_i) \}_{i=1}^{N_t}$ (where $N_t$ is a small number and label $\{y^t_i\}_{i=1}^{N_t}$ is bonafide), we adopt domain adaptation by mixing high frequency components from source domain with low frequency ones from target domain. To achieve this, we design a binary low frequency filter $\mathcal{F}_{low}$  to extract style contents from target domain, while a reversed high frequency filter $\mathcal{F}_{high}$ decompose bonafide/attack-related contents from source domain: 

\vspace{-0.5cm}
\begin{align}\label{filter}
&\mathbf{x}_{low }^t = \mathcal{F}_{low}(\mathcal{D}({\mathbf{x}^t}) ) , \\
&\mathbf{x}_{high}^s = \mathcal{F}_{high}(\mathcal{D}({\mathbf{x}^s}) ) , 
\end{align}
\vspace{-0.5cm}

Then, we apply frequency-base mixing to the obtained $\mathbf{x}_{low }^t$ and $\mathbf{x}_{high}^s$. The mixing is conducted in frequency domain and finally converted back to spatial domain via inverse DCT:

\vspace{-0.3cm}
 \begin{equation}\label{mix}
\mathbf{x}^{s \rightarrow t} =  \mathcal{D}^{-1}(\mathbf{x}_{low }^t + \mathbf{x}_{high}^s)
\vspace{-0.1cm}
\end{equation}

As bonafide/attack-related contents are contained in high frequency components from source domain, the mixed images should have identical labels as source images. 
We set low-frequency band to 2.5\% in our experiments. The mixed samples are shown in Fig. \ref{fig:example}.

% sota
\begin{table*}[t!]
 \begin{center}
\footnotesize
\caption{Comparison to existing SoTA methods on LivDet-Iris 2017 dataset under cross-dataset settings. }
\label{tab:sota}
  \begin{tabular}{c|cc|cc|cc}
  \toprule[2pt]
  Trained Dataset     & \multicolumn{2}{c|}{IIITD-WVU}    & \multicolumn{2}{c|}{NotreDame}    & \multicolumn{2}{c}{Clarkson}\\ \cline{1-7}
  Tested Dataset      & NotreDame       & Clarkson        & IIITD-WVU       & Clarkson        & IIITD-WVU       & NotreDame  \\
  \midrule[1.3pt]
  PBS\cite{fang2021iris}                 & \underline{16.86}          & 47.17          & 17.49          & 45.31          & 42.48          & 32.42         \\
  A-PBS\cite{fang2021iris}               & 27.61          & \textbf{21.99} & \textbf{9.49}  & \underline{22.46}          & \underline{34.17}          & \underline{23.08}         \\
  FAM+FMM(Ours)         & \textbf{5.81}           & \underline{26.03}          & \underline{15.07}          & \textbf{10.51}          & \textbf{22.06} & \textbf{20.92}\\
  \bottomrule[2pt]
 \end{tabular}
 \end{center}
\vspace{-0.8cm}
\end{table*}

%single domain
\begin{table*}[t!]
 \begin{center}
\footnotesize
\caption{Comparison to existing SoTA methods on LivDet-Iris 2017 dataset under intra-dataset settings. }
\label{tab:single}
 \begin{tabular}{c|c|ccccccc}
  \toprule[2pt]
  \multirow{2}{*}{Database}  & \multirow{2}{*}{Metric} & \multicolumn{7}{c}{PAD Algorithm(\%)}                                                                                                                                                                       \\ \cline{3-9} 
  &        & \multicolumn{1}{c|}{CASIA\cite{yambay2017livdet}} & \multicolumn{1}{c|}{SpoofNet\cite{kimura2020cnn}} & \multicolumn{1}{c|}{D-NetPAD\cite{sharma2020d}} & \multicolumn{1}{c|}{MSA\cite{fang2020micro}}   & \multicolumn{1}{c|}{PBS\cite{fang2021iris}}  & \multicolumn{1}{c|}{A-PBS\cite{fang2021iris}} & FAM \\ 
  \midrule[1.3pt]
  \multirow{3}{*}{Clarkson}  & APCER  & \multicolumn{1}{c|}{13.39}  & \multicolumn{1}{c|}{33.00}    & \multicolumn{1}{c|}{5.78}     & \multicolumn{1}{c|}{-}     & \multicolumn{1}{c|}{8.97} & \multicolumn{1}{c|}{6.16}  & 6.10                    \\
  & BPCER  & \multicolumn{1}{c|}{0.81}   & \multicolumn{1}{c|}{0.00}     & \multicolumn{1}{c|}{0.94}     & \multicolumn{1}{c|}{-}     & \multicolumn{1}{c|}{0.00} & \multicolumn{1}{c|}{0.81}  & 0.81                    \\
  & HTER   & \multicolumn{1}{c|}{7.10}   & \multicolumn{1}{c|}{16.50}    & \multicolumn{1}{c|}{\textbf{3.36}}     & \multicolumn{1}{c|}{-}     & \multicolumn{1}{c|}{4.48} & \multicolumn{1}{c|}{3.48}  & \underline{3.45}                    \\ \hline
  \multirow{3}{*}{NotreDame} & APCER  & \multicolumn{1}{c|}{7.78}   & \multicolumn{1}{c|}{18.05}    & \multicolumn{1}{c|}{10.38}    & \multicolumn{1}{c|}{12.28} & \multicolumn{1}{c|}{8.89} & \multicolumn{1}{c|}{7.88}     & 8.06                    \\
  & BPCER  & \multicolumn{1}{c|}{0.28}   & \multicolumn{1}{c|}{0.94}     & \multicolumn{1}{c|}{3.32}     & \multicolumn{1}{c|}{0.17}  & \multicolumn{1}{c|}{1.06} & \multicolumn{1}{c|}{0.00}  & 0.00                    \\
  & HTER   & \multicolumn{1}{c|}{\underline{4.03}}   & \multicolumn{1}{c|}{9.50}     & \multicolumn{1}{c|}{6.81}     & \multicolumn{1}{c|}{6.23}  & \multicolumn{1}{c|}{4.97} & \multicolumn{1}{c|}{\textbf{3.94}}  & \underline{4.03}                    \\ \hline
  \multirow{3}{*}{IIIT-WVU} & APCER  & \multicolumn{1}{c|}{29.40}   & \multicolumn{1}{c|}{0.34}     & \multicolumn{1}{c|}{36.41}    & \multicolumn{1}{c|}{2.31}  & \multicolumn{1}{c|}{5.76} & \multicolumn{1}{c|}{8.86}  & 1.00                    \\
  & BPCER  & \multicolumn{1}{c|}{3.99}   & \multicolumn{1}{c|}{36.89}    & \multicolumn{1}{c|}{10.12}    & \multicolumn{1}{c|}{19.94} & \multicolumn{1}{c|}{8.26} & \multicolumn{1}{c|}{4.13}  & 12.68                   \\
  & HTER   & \multicolumn{1}{c|}{16.70}   & \multicolumn{1}{c|}{18.62}    & \multicolumn{1}{c|}{23.27}    & \multicolumn{1}{c|}{11.13} & \multicolumn{1}{c|}{7.01} & \multicolumn{1}{c|}{\textbf{6.50}}  & \underline{6.84}                    \\ 
  \bottomrule[2pt]
 \end{tabular}
 \end{center}
\vspace{-0.7cm}
\end{table*}

\section{Experiments}
\label{sec:exp}

\subsection{Experimental Setup}
\label{ssec:setup}
\textbf{Datasets.}  The proposed method is evaluated on the LivDet-Iris 2017 dataset\cite{yambay2017livdet}, which consists of 4 different datasets. However, the Warsaw dataset is no longer publicly available, so we use the remaining Clarkson, Notre Dame, and IIITD-WVU datasets for evaluation. Metrics Attack Presentation Classification Error Rate (APCER), Bonafide Presentation Classification Error Rate (BPCER), and Half Total Error Rate (HTER) are deployed to measure the performance. 

\textbf{Implementation Details.}  The input iris image is grayscale and of 200 $\times$ 200 size. Random cropping is performed in the training phase. We employ ResNet18 as backbone and initial it with ImageNet pre-trained model. 
In domain adaptation task, the number of bonafide samples $N_t$ from target domain is 10, and source samples are randomly replaced with frequency-mixing samples by $p=0.5$. All the experiments are based on PyTorch.

\subsection{Ablation Studty}
\label{ssec:ablation}
In this section, we conduct ablation experiments to verify the effectiveness of frequency-based components in our proposed domain adaptation framework. 

\begin{figure}[t!]
\begin{center}
\setlength{\abovecaptionskip}{-1cm}
\includegraphics[width=0.9\linewidth]{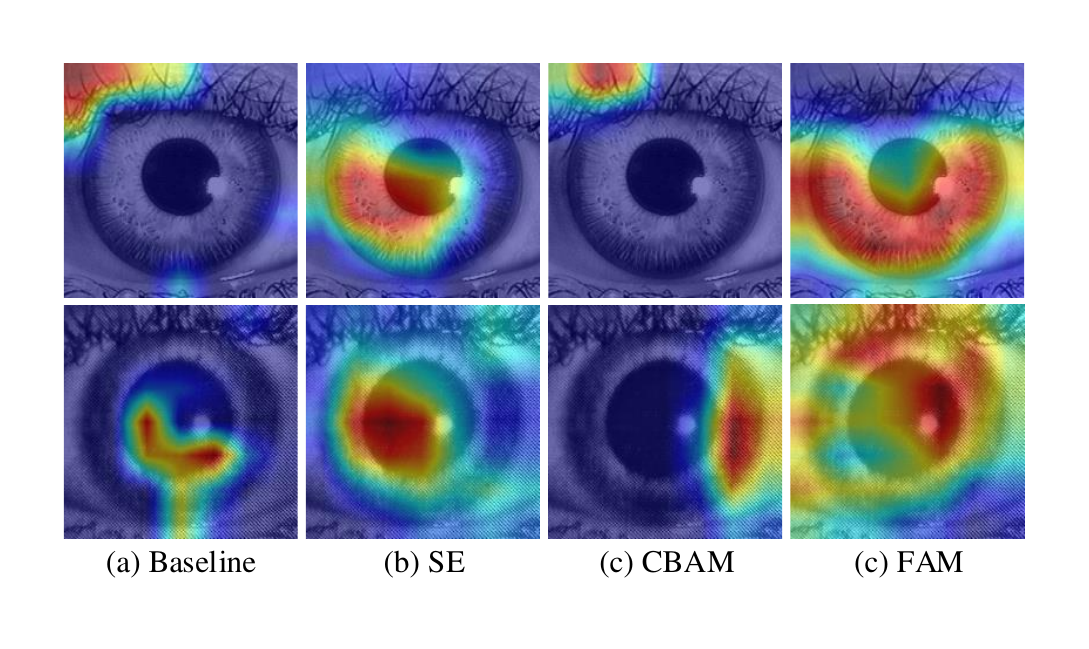}
\end{center}
\vspace{-1.2cm}
\caption{Grad-CAM\cite{selvaraju2017grad} visualization results. The first row is colored contact lens sample. The second row is printout attack sample. }
\label{fig:cam}
\vspace{-0.6cm}
\end{figure}

\textbf{Effect of frequency-based attention.} To testify the effectiveness of FAM where high frequency components are explicitly emphasized, we compare FAM with SE\cite{hu2018squeeze} and CBAM\cite{woo2018cbam}. Experiments are conducted both without domain adaptation (donated as wo/DA) and with domain adaptation (w/DA). As shown in Table \ref{tab:ablation}, average performance of FAM is constantly superior to SE and CBAM, especially under w/DA scenario. 
We apply the Grad-CAM\cite{selvaraju2017grad} to visualize class activation maps of attack class (see Fig. \ref{fig:cam}). Compared with other methods, it is clear that our FAM shows more interpretable attention and focuses on discriminative high-frequency regions according to different attacks. 
The improvement of FAM demonstrates the generalization ability of high frequency components. 

\textbf{Effect of frequency mixing domain adaptation.} FMM is designed for few-shot domain adaptation. To verify the effectiveness of its adaptation ability, we compare it with widely-used domain adaptation methods DANN\cite{ganin2016domain} and MMD\cite{tzeng2014deep}. Following the papers, we used the entire unlabeled target images for adaptation, while our FMM had limited access to only 10 bonafide images from target domain. Although DANN and MMD gain great improvement in certain settings, FMM still performs best considering the overall HTER. Adaptation results of FMM illustrate its efficacy even with few shots. Further integration of FAM, also referred to as our full FODA framework, increases average HTER to 16.73\%, improving 2.54\%. Thus, frequency mixing in our adaptation framework really helps to extract more adaptive features across datasets. 

\subsection{Comparison with State-of-the-Art Methods}
\label{ssec:sota}
We compare our method with current state-of-the-art methods in Table \ref{tab:sota}. The results show that our method performs better than PBS\cite{fang2021iris} consistently and surpasses A-PBS\cite{fang2021iris} in the majority of settings. 
It is worth noting that Clarkson dataset has a quite distinct style compared to the others, and we achieve 11.95\%, 12.11\% improvement when adapting from NotreDame to Clarkson, Clarkson to IIITD-WVU respectively. Our method integrates source bonafide/attack-related components to style contents from target domain and boosts the performance with very few costs. 

As few methods in the literature focus on cross-dataset iris PAD, Table \ref{tab:single} also reports results of FAM under intra-dataset settings. We still observe superior or comparable results compared to SoTA methods. It confirms high frequency components are effective in iris presentation attack detection.

\section{Conclusion}
\label{sec:conclu}
In this paper, we propose a novel few-shot one-class domain adaptation framework, including frequency-based attention module (FAM) for high frequency component highlighting and frequency mixing module (FMM) for style-adapted domain adaptation. FAM aggregates frequency information into spatial attention and emphasizes fine-grained features. FMM mixes low frequency components of source images with high frequency components of very few target bonafide images. Extensive experiments show the effectiveness of the proposed method under both cross-dataset and intra-dataset settings.

\vfill\pagebreak

\bibliographystyle{IEEEbib}
\bibliography{refs}

\end{document}